\documentclass[preprint,12pt,sort&compress]{elsarticle}

\usepackage{textcomp,booktabs}
\usepackage[usenames,dvipsnames]{color}
\usepackage{colortbl}
\definecolor{mygray}{gray}{.9}
\definecolor{mypink}{rgb}{.99,.91,.95}
\definecolor{mycyan}{cmyk}{.3,0,0,0}
\usepackage{amssymb}
\usepackage{graphicx}
\usepackage{booktabs}
\usepackage{tabularx}
\usepackage{array}
\usepackage{lscape}
\usepackage{longtable}
\usepackage{multirow}
\usepackage{amsmath}
\usepackage{colortbl}
\usepackage[figuresright]{rotating}
\usepackage{epsfig}
\usepackage{epsf}
\usepackage{subfigure}
\usepackage{threeparttable}

\newcommand{\PreserveBackslash}[1]{\let\temp=\\#1\let\\=\temp}
\newcolumntype{C}[1]{>{\PreserveBackslash\centering}p{#1}}
\newcolumntype{R}[1]{>{\PreserveBackslash\raggedleft}p{#1}}
\newcolumntype{L}[1]{>{\PreserveBackslash\raggedright}p{#1}}

\journal{} 
\begin{document}

\begin{frontmatter}



\title{An evidential Markov decision making model}


\author[address1]{Zichang He}
\author[address1]{Wen Jiang\corref{label1}}
\address[address1]{School of Electronics and Information, Northwestern Polytechnical University, Xi'an, Shaanxi, 710072, China}
\cortext[label1]{Corresponding author at Wen Jiang: School of Electronics and Information, Northwestern Polytechnical University, Xi'an, Shaanxi 710072, China. Tel: (86-29)88431267. E-mail address: jiangwen@nwpu.edu.cn, jiangwenpaper@hotmail.com}

\begin{abstract}
The sure thing principle and the law of total probability are basic laws in classic probability theory. A disjunction fallacy leads to the violation of these two classical laws. In this paper, an Evidential Markov (EM) decision making model based on  Dempster-Shafer (D-S) evidence theory and Markov modelling is proposed to address this issue and model the real human decision-making process. In an evidential framework, the states are extended by introducing an uncertain state which represents the hesitance of a decision maker. The classical Markov model can not produce the disjunction effect, which assumes that a decision has to be certain at one time. However, the state is allowed to be uncertain in the EM model before the final decision is made. An extra uncertainty degree parameter is defined by a belief entropy, named Deng entropy, to assignment the basic probability assignment of the uncertain state, which is the key to predict the disjunction effect. A classical categorization decision-making experiment is used to illustrate the effectiveness and validity of EM model. The disjunction effect can be well predicted and the free parameters are less compared with the existing models.
\end{abstract}

\end{frontmatter}

\section{Introduction}\label{Introduction}
The sure thing principle introduced by Jim Savage\cite{Jarrett1956The} is fundamental in economics and probability theory. It means that if one prefers action $A$ over $B$ under state of the world $X$, while action $A$ is also preferred under the complementary state $\neg X$, then it can be concluded that one will still prefer action $A$ over $B$ under the state is unknown.
The law of total probability is a fundamental rule of Bayesian probability relating marginal probabilities to conditional probabilities.
It expresses the total probability of an outcome which can be realized via several distinct events.
However, many experiments and studies have shown that the sure thing principle and the law of total probability can be violated due to the disjunction effect\cite{Lambdin2007The,Townsend2000Exploring,pothos2009quantum}.
The disjunction fallacy is an empirical finding in which the proportion taking the target gamble under the unknown condition falls below both of the proportions taking the target gamble under each of the known conditions.
The same person takes the target gamble under both known conditions, but then rejects the target gamble under the unknown condition\cite{Tversky1992The}.

To explain the disjunction fallacy, many studies have been proposed.
The original explanation was a psychological idea based on the failure of consequential reasoning under unknown condition\cite{Tversky1992The}.
A Markov decision making model was proposed by Townsend $et\ al.$ (2000)\cite{Townsend2000Exploring}.
A Markov process can be used to model a random system that changes states according to a transition rule that only depends on the current state.
Markov models has been used in many applications\cite{Annett2010Twelve,Alagoz2010Markov,Farahat2010Markov,Navaei2010MARKOV},  especially in the prediction\cite{Lu2014Application,Samet2014Enhancement}, such as rainfall prediction\cite{fraedrich1983single,Liu2011Study}, economy prediction\cite{jin2015phev,mar2010probabilistic} and so on.
However, it failed to predict the disjunction effect, which will be introduced later.
More recently, the theory of quantum probability has been introduced in the cognition and decision making process.
Quantum probability is an effective approach to psychology and behavioristics \cite{wang2013potential,pothos2015structured,aerts2013concepts,blutner2013quantum}.
It has been widely applied to the fields of cognition and decision making fields to explain the phenomena in classical theory, like order effect\cite{trueblood2011quantum,wang2013quantum,wang2014context}, disjunction effect\cite{pothos2009quantum}, the interference effect of categorization\cite{wang2016interference}, prisoner's dilemma\cite{chen2003quantum}, conceptual combinations\cite{bruza2015probabilistic,aerts2009quantum}, quantum game theory\cite{Situ2016Relativistic,Situ2016Two,Asano2010Quantum} and so on.
To explain the disjunction fallacy, many quantum models have been proposed, such as a quantum dynamical (QD) model\cite{Pothos2009A,Busemeyer2009Empirical}, quantum-like models\cite{Denolf2016Bohr,Nyman2011On,Nyman2011Quantum}
quantum prospect decision theory\cite{Yukalov2015Quantum,Yukalov2009Physics,Yukalov2009Processing,Yukalov2011Decision} and quantum-like Bayesian networks\cite{Moreira2016Quantum} etc.
In a quantum framework, the decision in human brain is deemed as a superposition of several decisions before the final one is made. Although the quantum model works for explaining the disjunction fallacy, the major problem is the additional introduction of quantum parameters.

Decision making and optimization under uncertain environment is normal in reality and is heavily studied\cite{Jiang2016sensor,Jiang2016A,dengentropy}.
It is still an open issue for uncertain information modeling and processing\cite{Tang2017A,Deng2016Evidence,Jiang2017mGCR}.
Some methods have been proposed to handle the uncertainty like probability theory\cite{Frank1967An}, fuzzy set theory\cite{Zadeh1965Fuzzy} , Dempster-Shafer evidence theory\cite{Dempster1967Upper,Shafer1978A}, rough sets\cite{Pawlak1982Rough}, and D-numbers\cite{deng2012d}.
Dempster-Shafer (D-S) evidence theory\cite{Dempster1967Upper,Shafer1978A} is a powerful tool to handle the uncertainty. It has been widely used in many fields, like risk analysis\cite{Fu2015A,Su2015Dependence,Jiang2017FMEA}, controller design\cite{Tang2016A,Yager1995Including}, pattern recognition\cite{Han2016A,Ma2016An,Liu2016Adaptive}, fault diagnosis\cite{Yuan2016Conflict,Yuan2016Modeling,Jiang2016An}, multiple attribute decision making\cite{Fu2015An,Du2016Attribute} and so on. Also D-S theory is widely combined with Markov models\cite{Zhou2013A,Deng2015Newborns,Soubaras2009An,Soubaras2010On}.

In this paper, an evidential Markov (EM) decision making model based on D-S evidence theory and Makov modelling is proposed to model the decision making process.
In the paradigm of studying the disjunction fallacy, generally a decision making process is consisted of beliefs and actions.
The EM model assumes that people's decision can be uncertain before the final one is made while a it is assumed that people is decision has to be certain at one time.
In an evidential framework, the actions states are extended by introducing an uncertain state to represent the hesitance of a decision maker.
An extra uncertainty degree is determined by a belief entropy, named Deng entropy, which is the key to predict the disjunction effect in our model.
The EM model is used in an categorization decision-making experiment.
The model results is discussed and compared, which show the effectiveness and validity of our model.

The rest of the paper is organized as follows. In Section 2, the preliminaries of basic theories employed are briefly introduced. The classical Markov decision making model is introduced in Section 3. Then a categorization decision-making experiment is illustrated in Section 4. Our EM model is proposed and applied to explain the experiment result in Section 5. The model result is discussed and compared in Section 6. Finally, Section 7 comes to the conclusion.
\section{Preliminaries}
\subsection{Dempster-Shafer evidence theory}
In D-S theory, a finite set of $N$ mutually exclusive and exhaustive elements, called the frame of discernment (FOD), which is symbolized by $\Theta {\rm{ = }}\left\{ {{H_1},{H_2}, \ldots ,{H_N}} \right\}$. Let us denote $P\left( \Theta  \right)$ as the power set composed of ${2^N}$ elements $A$ of $\Theta$:
\[P\left( \Theta  \right) = \left\{ {\emptyset ,\left\{ {{H_1}} \right\},\left\{ {{H_2}} \right\}, \ldots ,\left\{ {{H_N}} \right\},\left\{ {{H_1} \cup {H_2}} \right\},\left\{ {{H_1} \cup {H_3}} \right\}, \ldots ,\Theta } \right\}\]
A basic probability assignment (BPA) is a mapping from $P\left( \Theta  \right)$ to $\left[ {0,1} \right]$, defined as $m: P\left( \Theta  \right) \to \left[ {0,1} \right]$ satisfying\cite{Dempster1967Upper,Shafer1978A}:
\begin{equation}
\begin{array}{l}
\sum\limits_{A \in P\left( \Theta  \right)} {m\left( A \right)}  = 1,\\
m\left( \emptyset  \right) = 0.
\end{array}
\end{equation}
The mass function $m$ represents supporting degree to focal element $A$.
The elements of $P\left( \Theta  \right)$ that have a non-zero mass are called focal elements.
A body of evidence (BOE) is the set of all the focal elements, defined as
\[\left( {\Re ,m} \right) = \left\{ {\begin{array}{*{20}{c}}
{\left[ {A,m\left( A \right)} \right];}&{A \in P\left( \Theta  \right),m > 0}
\end{array}} \right\},\]
A mass function corresponds to a belief ($Bel$) function and a plausibility ($Pl$) function respectively.
Given $m:{P\left( \Theta  \right)} \to \left[ {0,1} \right]$, $Bel\left( A \right)$ function represents the whole belief degree to $A$, defined as
\begin{equation}
\begin{array}{*{20}{c}}
{Bel\left( A \right) = \sum\limits_{B \subseteq A} {m\left( B \right);} }&{\forall A \subseteq {P\left( \Theta  \right)}}
\end{array}
\end{equation}
$Pl$ function represents the belief degree of not denying $A$, defined as
\begin{equation}
\begin{array}{*{20}{c}}
{Pl\left( A \right) = \sum\limits_{B \cap A \ne \emptyset } {m\left( B \right);} }&{\forall A \subseteq {P\left( \Theta  \right)}}
\end{array}
\end{equation}
It should be noticed that when all the focal sets of $m$ are singletons, $m$ is said to be $Bayesian$; $Bel$ and $Pl$ then degenerate into the same probability measure.

\begin{figure}[!ht]
\centering
\includegraphics[scale=0.65]{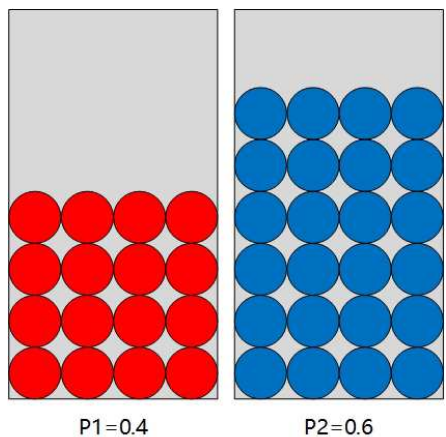}
\caption{A game of picking ball which can be handled by probability theory}\label{randb}
\end{figure}

\begin{figure}[!ht]
\centering
\includegraphics[scale=0.65]{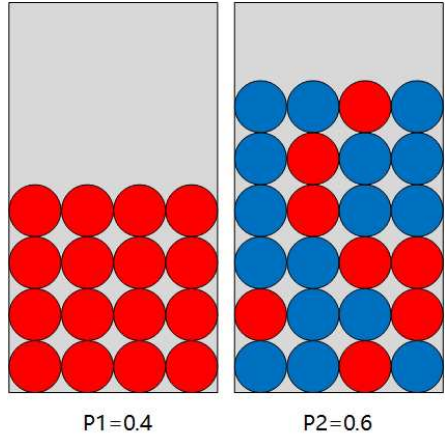}
\caption{A game of picking ball where probability theory is unable but D-S evidence theory is able to handle}\label{randrb}
\end{figure}

In the following, an game of picking ball will be used to show the D-S theory's ability of handling uncertainty\cite{dengentropy}. There are two boxes filled with some balls as shown in Figure \ref{randb}. Left box is contended with red balls and right box is contended with blue balls. The number of balls in each box is unknown. Now, a ball is picked randomly from two boxes. The probability of picking from left box $P1$ is known as 0.4 while picking from right box $P2$ is known as 0.6. It is easy to obtain that the probability of picking a red ball is 0.4 while picking a blue ball is 0.6 based on probability theory.
Now, the situation changes as shown in Figure \ref{randrb}. The left box is contended with right balls while the right box is contended with red and blue balls. The exact number of the balls in each box and the ratio of red balls with blue balls are completely unknown. The probabilities of selecting from two boxes keep the same, $P1 = 0.4$ and $P2 = 0.6$. The question is what the probability that a red ball is picked is. Due to the lack of information, the question can not be addressed in probability theory. However, D-S evidence theory can effectively handle it. We can obtain two mass functions that $m\left( R \right) = 0.4$ and $m\left( {R,B} \right) = 0.6$. Then the uncertainty is well handled in the frame of D-S theory.

\subsection{Pignistic probability transformation}
The term "pignistic" proposed by Smets is originated from the word pignus, meaning bet in Latin. Pignistic probability transformation (PPT) has a wide application in decision making. Principle of insufficient reason is used to assign the probability of a multiple-element set to singleton sets. In other word, a belief interval is distributed into the crisp one determined as\cite{Smets1994The}:
\begin{equation}{\label{PPT}
Bet\left( {A} \right) = \sum\nolimits_{A \subseteq B} {\frac{{m\left( {B} \right)}}{{\left| {B} \right|}}}
}\end{equation}
where $m$ is a mass function, $B$ is the focal element of $m$ and ${\left| {B} \right|}$ is the cardinality of $B$ which denotes the number of elements in set $B$.

\subsection{Deng entropy}
Deng entropy is proposed to measure uncertainty, which is defined as follows\cite{dengentropy}:
\begin{equation}\label{Dengentropy}
{E_d} =  - \sum\limits_{A \in X} {m\left( A \right){{\log }_2}\frac{{m\left( A \right)}}{{{2^{\left| A \right|}} - 1}}},
\end{equation}
where $m$ is a mass function defined on FOD $X$, $A$ is the focal element of $m$ and ${\left| A \right|}$ is the cardinality of $A$.
Especially, for a BOE $\left( {\Re ,m} \right)$, if all the focal elements are singletons, Deng entropy degenerates into Shannon entropy.
\begin{equation}
{E_d} =  - \sum\limits_{A \in X} {m\left( A \right){{\log }_2}\frac{{m\left( A \right)}}{{{2^1} - 1}}} {\rm{ = }} - \sum\limits_{A \in X} {m\left( A \right){{\log }_2}m\left( A \right)}
\end{equation}
\section{Classical Markov decision-making model}
Townsend $et\ al.$ (2000)\cite{Townsend2000Exploring} originally proposed a Markov model for studying the disjunction effect. For simplicity, we use a two-dimensional Markov model to illustrate the decision making process.
\subsection{State representation}
\begin{figure}[!ht]
\centering
\includegraphics[scale=0.65]{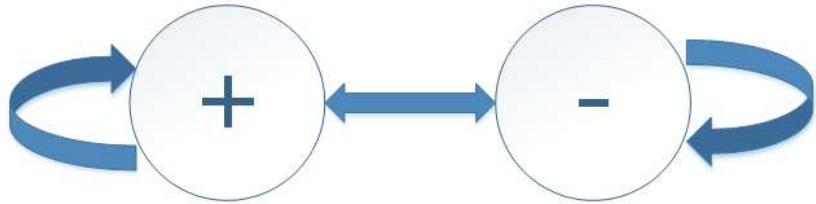}
\caption{Two-state transition digram}\label{twoM}
\end{figure}

The model assumes that the perceptual system can be in one of two states at each moment in time: a plus state denoted $\left| {\rm{ + }} \right\rangle $ or a minus state denoted $\left| {\rm{ - }} \right\rangle $ representing the other orientation. A sample path by the Markov process starts in some initial state, either $\left| {S\left( 0 \right)} \right\rangle {\rm{ = }}\left| {\rm{ + }} \right\rangle $ or $\left| {S\left( 0 \right)} \right\rangle {\rm{ = }}\left| {\rm{ - }} \right\rangle $, and jumps back and forth across time from a clear plus state to a clear minus state (Figure \ref{twoM}).
In general, however, we do not know which it is for any given sample path, and so the uncertainty is represented by assigning an initial probability distribution across these two possibilities, denoted ${\phi _{\rm{ + }}}\left( 0 \right)$ for plus and ${\phi _{\rm{ - }}}\left( 0 \right)$ for minus, and ${\phi _{\rm{ + }}}\left( 0 \right){\rm{ + }}{\phi _{\rm{ - }}}\left( 0 \right){\rm{ = }}0$.
\subsection{State transition}
The initial state is defined by a probability distribution represented by a $2 \times 1$ matrix
\[\phi \left( 0 \right){\rm{ = }}\left[ {\begin{array}{*{20}{c}}
{{\phi _{\rm{ + }}}\left( 0 \right)}\\
{{\phi _{\rm{ - }}}\left( 0 \right)}
\end{array}} \right].\]
If the system is known to start from the plus state, then ${\phi _{\rm{ + }}}\left( 0 \right) = 1$; if the state is known to start from the minus state, then ${\phi _{\rm{ - }}}\left( 0 \right) = 1$; and if the state is unknown, then ${\phi _{\rm{ + }}}\left( 0 \right) + {\phi _ - }\left( 0 \right) = 1$. After some period of time ${t_1}$, the perceptual system can transit to a new state. Hence, a probability transition matrix is determined as
\[T\left( {{t_1},0} \right) = \left[ {\begin{array}{*{20}{c}}
{{T_{ + , + }}\left( {{t_1},0} \right)}&{{T_{ + , - }}\left( {{t_1},0} \right)}\\
{{T_{ - , + }}\left( {{t_1},0} \right)}&{{T_{ - , - }}\left( {{t_1},0} \right)}
\end{array}} \right],\]
where, for example, ${{T_{ + , - }}\left( {{t_1},0} \right)}$ represents the probability of transferring from the minus state at time zero to the plus state at time ${t_1}$. It is called a stochastic matrix because all the elements are nonnegative and sum to unity with a column.
The matrix product of the transition matrix times the initial probability distribution at time zero produces the updated distribution across the two states at time ${t_1}$:
\[\phi \left( {{t_1}} \right) = T\left( {{t_1},0} \right) \cdot \phi \left( 0 \right).\]
Given the state at time ${t_1}$, we can update again using the transition matrix $T\left( {{t_2},{t_1}} \right)$ that describes the probabilities of transitions for the period of time from ${t_1}$ to ${t_2}$:
\[\phi \left( {{t_2}} \right) = T\left( {{t_2},{t_1}} \right) \cdot \phi \left( {{t_1}} \right) = T\left( {{t_2},{t_1}} \right) \cdot T\left( {{t_1},0} \right) \cdot \phi \left( 0 \right).\]
It is reasonable to assume that the transition matrix remains stationary for many applications. Hence, the transition matrix can be simply write as $T\left( t \right)$, in which $t$ is the duration time period. In this case, the probability distribution at ${t_2}$ equals
\begin{equation}
\phi \left( {{t_2}} \right) = T\left( {{t_2}} \right) \cdot \phi \left( 0 \right) = T\left( {{t_2} - {t_1}} \right) \cdot T\left( {{t_1}} \right) \cdot \phi \left( 0 \right).
\end{equation}
It means that the transition matrix obeys the semi-group property of dynamic systems,
\[T\left( {t + u} \right) = T\left( t \right) \cdot T\left( u \right) = T\left( u \right) \cdot T\left( t \right).\]
The time evolution of the transition matrix should obey the following differential equation, called the Kolmogorov forward equation.
\begin{equation}
\frac{{dT\left( t \right)}}{{{\rm{d}}t}} = K \cdot T\left( {\rm{t}} \right)
\end{equation}
Also the time evolution of the probability distribution over states should obey:
\begin{equation}\label{Kolmogorov}
\frac{{d\phi \left( t \right)}}{{{\rm{d}}t}} = K \cdot \phi \left( {\rm{t}} \right).
\end{equation}
The ${2 \times 2}$ matrix $K$ is called the intensity matrix, which is used to construct the transition matrix as a function of processing time duration. The matrix must satisfy the following constraints to guarantee that the solution is a transition matrix: the off-diagonal elements must be nonnegative but the sum to zero with a column.
The solution of Kolmogorov equation is a matrix exponential function:
\begin{equation}\label{KolSolution}
\begin{array}{l}
T\left( t \right) = {e^{tK}},\\
\phi \left( t \right) = {e^{tK}} \cdot \phi \left( 0 \right).
\end{array}
\end{equation}
\subsection{Response probabilities}\label{responsepro}
A response refers to a measurement that the perceiver maker or reports. To derive response probabilities at various time points from the Markov model, measure matrixes are defined as
\[\begin{array}{*{20}{c}}
{{M_ + } = \left[ {\begin{array}{*{20}{c}}
1&0\\
0&0
\end{array}} \right]and}&{{M_ - } = \left[ {\begin{array}{*{20}{c}}
0&0\\
0&1
\end{array}} \right]},
\end{array}\]
where ${M_+}$ is applied to measure the plus state and ${M_-}$ is applied to measure the minus state. Let $R\left( t \right) =  + $ denote the response that the system is in the plus state at time $t$. In the following, it will be convenient to use the $1 \times 2$ matrix $L = \left[ {1,1} \right]$ perform summation across states. Using these matrixes, the probability if observing a "plus" at time $t$ equals
\begin{equation}
p\left( {R\left( t \right) =  + } \right) = L \cdot {M_ + } \cdot T\left( t \right) \cdot \phi \left( 0 \right).
\end{equation}
\subsection{The law of total probability}\label{LawinM}
Assume that the initial state is known to "plus" or "minus", the probability of $R\left( t \right) =  + $ equals
\[\begin{array}{l}
p\left( {R\left( t \right) =  + | + } \right) = \left[ {1,1} \right] \cdot \left[ {\begin{array}{*{20}{c}}
1&0\\
0&0
\end{array}} \right] \cdot \left[ {\begin{array}{*{20}{c}}
{{T_{ + , + }}\left( t \right)}&{{T_{ + , - }}\left( t \right)}\\
{{T_{ - , + }}\left( t \right)}&{{T_{ - , - }}\left( t \right)}
\end{array}} \right]\left[ {\begin{array}{*{20}{c}}
1\\
0
\end{array}} \right] = {T_{ + , + }}\left( t \right),\\
p\left( {R\left( t \right) =  + | - } \right) = \left[ {1,1} \right] \cdot \left[ {\begin{array}{*{20}{c}}
1&0\\
0&0
\end{array}} \right] \cdot \left[ {\begin{array}{*{20}{c}}
{{T_{ + , + }}\left( t \right)}&{{T_{ + , - }}\left( t \right)}\\
{{T_{ - , + }}\left( t \right)}&{{T_{ - , - }}\left( t \right)}
\end{array}} \right]\left[ {\begin{array}{*{20}{c}}
0\\
1
\end{array}} \right] = {T_{ + , - }}\left( t \right)
\end{array}\]
respectively. Assume that the initial state is unknown, then the probability of $R\left( t \right) =  + $ equals
\[\begin{array}{l}
p\left( {R\left( t \right) =  + |U} \right) = \left[ {1,1} \right] \cdot \left[ {\begin{array}{*{20}{c}}
1&0\\
0&0
\end{array}} \right] \cdot \left[ {\begin{array}{*{20}{c}}
{{T_{ + , + }}\left( t \right)}&{{T_{ + , - }}\left( t \right)}\\
{{T_{ - , + }}\left( t \right)}&{{T_{ - , - }}\left( t \right)}
\end{array}} \right]\left[ {\begin{array}{*{20}{c}}
{{\phi _{\rm{ + }}}\left( 0 \right)}\\
{{\phi _{\rm{ - }}}\left( 0 \right)}
\end{array}} \right]\\
 = {T_{ + , + }}\left( t \right){\phi _{\rm{ + }}}\left( 0 \right) + {T_{ + , - }}\left( t \right){\phi _{\rm{ - }}}\left( 0 \right).
\end{array}\]
The last line expresses that the probability of being the plus state for the unknown condition equals to the probability of reaching it by two different paths (see Figure \ref{+M}), which means that the law of total probability is not violated.

Hence, the classical Markov model can not explain and predict the disjunction effect. To address it, an evidential Markov model is proposed. There are several paradigms for studying the disjunction fallacy, in the following, a categorization decision-making experiment is briefly introduced.
\section{Categorization decision-making experiment}
\subsection{Experiment method}
\begin{figure}[!ht]
\centering
\includegraphics[scale=1]{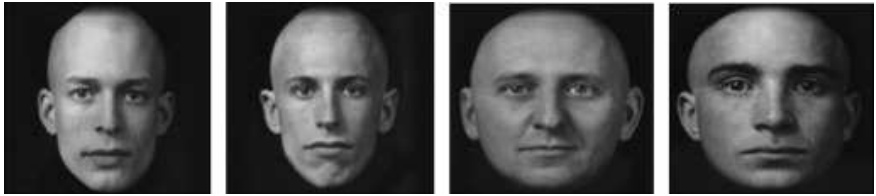}
\caption{Example faces used in a categorization decision-making experiment}\label{face}
\end{figure}

Townsend $et\ al.$\cite{Townsend2000Exploring} proposed a categorization decision-making experiment to study the interactions between categorization and decision making. It turned out that categorization can result in the disjunction effect on decision making.
In the experiment, pictures of faces varying along face width and lip thickness are shown to participants. Generally, the faces can be distributed into two different kinds: on average a narrow (N) face with a narrow width and thick lips; on average a wide (W) face with a wide width and thin lips (see Figure \ref{face} for example). The
The participants were informed that N face had a 0.60 probability to come from the "bad" guy population while W face had a 0.60 probability to come from the "good" guy population. The participants were usually (probability 0.70) rewarded for attacking "bad guys" and they were usually (probability 0.7) rewarded for withdrawing from "bad guys". The primary manipulation was produced by using the following two test conditions, presented across a series of trials, to each participant.
In the categorization-decision making (C-D) condition: participants were asked to categorize a face as belonging to either a "good" (G) guy or "bad" (B) guy group. Following the categorization, they were asked to make a decision whether to "attack" (A) or to to "withdraw" (W).
In the decision-making alone (D-alone) condition: participants were only asked to make an action decision.
The experiment included a total of 26 participants, but each participant provided 51 observations for the C-D condition for a total of ${\rm{26}} \times {\rm{51 = 1326}}$ observations, while each person produced 17 observations for the D condition for a total of ${\rm{17}} \times {\rm{26 = 442}}$ total observations.
\subsection{Experiment results}
The experiment results are shown in Table \ref{classical result}.
\begin{table}[!h]
\centering
\caption{The results of Townsend $et\ al.$\cite{Townsend2000Exploring}}
\label{classical result}
\begin{tabular}{cccccccc}
\toprule
Type face & \textbf{$P\left( {G} \right)$} & \textbf{$P\left( {A|G} \right)$} & \textbf{$P\left( {B} \right)$} & \textbf{$P\left( {A|B} \right)$} & \textbf{${P_T}$} & \textbf{$P\left( {A} \right)$}  \\
\midrule
Wide               & 0.84          & 0.35            & 0.16          & 0.52            & 0.37       & 0.39          \\
Narrow             & 0.17          & 0.41            & 0.83          & 0.63            & 0.59       & 0.69          \\
\bottomrule
\end{tabular}
\end{table}
The column labeled $P\left( {G} \right)$ shows the probability of categorizing the face as a "good buy";
the column labeled $P\left( {A|G} \right)$ shows the probability of attacking given categorizing the face as a "good guy";
the column labeled $P\left( {B} \right)$ shows the probability of categorizing the face as a "bad buy";
the column labeled $P\left( {A|B} \right)$ shows the probability of attacking given categorizing the face as a "bad guy";
and the column labeled ${P_T}$ shows the total probability of attacking in the C-D condition computed by
\begin{equation}\label{Pt}
P\left( A \right) = P(G) \cdot P\left( {A|G} \right) + P\left( B \right) \cdot P\left( {A|B} \right),
\end{equation}
and the column labeled $P\left( {A} \right)$ represents the probability of attacking in the D-alone condition.
Obviously, ${P_T}$ and $P\left( {A} \right)$ are unequal for both types of face.
According to Bayesian probability theory, this violates the law of total probability and the difference value is called the disjunction effect. In this experiment, the disjunction effect for wide type faces is so weak that it can be ignored and explained as the statistical error, however, the disjunction effect for narrow type is too prominent to ignore.

The classical paradigm has been discussed in many later works. Literatures of studying the categorization decision-making experiment and their results are shown below in Table \ref{preworks}.
\begin{table}[!h]
\centering
\small
\caption{Results of categorization decision-making experiments}
\label{preworks}
\begin{threeparttable}
\begin{tabular*}{\columnwidth}{ccccccccc}
\toprule
Literature                                                                                           & Type & $P\left( {G} \right)$ & $P\left( {A|G} \right)$ & $P\left( {B} \right)$ & $P\left( {A|B} \right)$ & ${P_T}$   & $P\left( {A} \right)$   \\
\midrule
\multirow{2}{*}{\scriptsize\begin{tabular}[c]{@{}c@{}}Townsend $et\ al.$ (2000)\cite{Townsend2000Exploring}\end{tabular}}                     & W    & 0.84 & 0.35   & 0.16 & 0.52   & 0.37 & 0.39  \\
                                                                                                     & N    & 0.17 & 0.41   & 0.83 & 0.63   & 0.59 & 0.69   \\ \hline
\multirow{2}{*}{\scriptsize\begin{tabular}[c]{@{}c@{}}Busemeyer $et\ al.$ (2009)\cite{Busemeyer2009Empirical}\end{tabular}}                    & W    & 0.80 & 0.37   & 0.20 & 0.53   & 0.40 & 0.39  \\
                                                                                                     & N    & 0.20 & 0.45   & 0.80 & 0.64   & 0.60 & 0.69   \\ \hline
\multirow{2}{*}{\scriptsize\begin{tabular}[c]{@{}c@{}}Wang \& Busemeyer (2016)\\  Experiment 1\cite{wang2016interference}\end{tabular}} & W    & 0.78 & 0.39   & 0.22 & 0.52   & 0.42 & 0.42      \\
                                                                                                     & N    & 0.21 & 0.41   & 0.79 & 0.58   & 0.54 & 0.59   \\ \hline
\multirow{2}{*}{\scriptsize\begin{tabular}[c]{@{}c@{}}Wang \& Busemeyer (2016) \\ Experiment 2\cite{wang2016interference}\end{tabular}} & W    & 0.78 & 0.33   & 0.22 & 0.53   & 0.37 & 0.37     \\
                                                                                                     & N    & 0.24 & 0.37   & 0.76 & 0.61   & 0.55 & 0.60   \\ \hline
\multirow{2}{*}{\scriptsize\begin{tabular}[c]{@{}c@{}}Wang \& Busemeyer (2016) \\ Experiment 3\cite{wang2016interference}\end{tabular}} & W    & 0.77 & 0.34   & 0.23 & 0.58   & 0.40 & 0.39  \\
                                                                                                     & N    & 0.24 & 0.33   & 0.76 & 0.66   & 0.58 & 0.62   \\\hline
\multirow{2}{*}{Average}                                                                             & W    & 0.79 & 0.36   & 0.21 & 0.54   & 0.39 & 0.39      \\
                                                                                                     & N    & 0.21 & 0.39   & 0.79 & 0.62   & 0.57 & 0.64 \\
\bottomrule
\end{tabular*}
 \begin{tablenotes}
        \footnotesize
        \item[1] In Busemeyer $et\ al.$ (2009)\cite{Busemeyer2009Empirical}, the classical experiment was replicated.
        \item[2] In Wang and Busemeyer (2016)\cite{wang2016interference}, experiment 1 used a larger data set to replicate the classical experiment. Experiment 2 introduced a new X-D trial verse C-D trial and only the results of C-D trial are used here. In experiment 3, the reward rate for attacking bad people was a bit less than experiment 1 and 2.
      \end{tablenotes}
\end{threeparttable}
\end{table}
\section{Evidential Markov model}\label{Proposed method}
The EM model based on D-S evidence theory and Markov modelling is proposed to model the human decision making process. In the following, we will use the categorization decision-making experiment to illustrate our model.
The disjunction fallacy can be explained and the disjunction effect for narrow type faces can be well predicted in our model.
The flowchart of EM model is shown in Figure \ref{flowchart}.

\textbf{Step 1: Frame of discrimination determined}

In the C-D condition, the outcome will be either to attack or withdraw given the face is categorized as a good guy or a bad guy. Thus the FOD ${\Theta _{CD}}$ is determined as ${\Theta _{CD}} = \left\{ {AG,WG,AB,WB} \right\}$, where, for example, $AG$ represents the participant decide to attack given the face is categorized as a good guy.
In this case, as the categorization is uncertain to the participant, it is meaningless to extend the belief states.
However, in an evidential framework, the action states should be extended.
Then the states $AWG$ (denoted $UG$), $AWB$ (denoted $UB$) are introduced to fill the power set of ${\Theta_{CD}}$. The state $UG$ represents that the participant is uncertain to attack or withdraw given the face is categorized as $G$; the state $UB$ represents that the participant is uncertain to attack or withdraw given the face is categorized as bad. In brief, the hesitance of a participant is shown by $UG$ and $UB$.

In the D-alone condition, the outcome of the game will be either to attack or withdraw without a precise categorization.
Thus the FOD ${\Theta _D}$ is determined as ${\Theta _D}  = \left\{ {AU,WU} \right\}$, where, for example, $AU$ represents that a participant decide to attack given the categorization is unknown. In this case, the action states should be extended in an evidential framework. But the state of belief is unknown, the belief state is actually an assembly of categorizing the face as good and bad.
Then the state $AWU$ (denoted $UU$) is introduced to fill the power set of ${\Theta _D}$. The state $UU$ represents that a participant is uncertain to attack or withdraw given that the categorization is unknown, which also shows the hesitance of a participant.

\textbf{Step 2: Representation of beliefs and actions}

In general, the beliefs and actions consist of the decision making process.
As the states have been extended, the initial state involves six combination of beliefs and actions
\[\left\{ {\left| {{B_G}{A_A}} \right\rangle ,\left| {{B_G}{A_U}} \right\rangle ,\left| {{B_G}{A_W}} \right\rangle ,\left| {{B_B}{A_A}} \right\rangle ,\left| {{B_B}{A_U}} \right\rangle ,\left| {{B_C}{A_W}} \right\rangle } \right\},\]
where, for example, ${\left| {{B_G}{A_A}} \right\rangle }$ symbolizes the event in which the player believes the face belongs to a good guy, but the player intends to act by attacking.
All of the possible transitions between the six states are illustrated in Figure \ref{sixstate}.

The Markov model assumes that a person states in exactly one circle, but we do not known which one is is. Hence, we assign probabilities of to the possible states. At the beginning of a trail, the initial probability of starting out in one of these six states is defined by a probability distribution represented by the $6 \times 1$ column matrix
\[\phi \left( {\rm{0}} \right){\rm{ = }}\left[ {\begin{array}{*{20}{c}}
{{\phi _{AG}}}\\
{{\phi _{UG}}}\\
{{\phi _{WG}}}\\
{{\phi _{AB}}}\\
{{\phi _{UB}}}\\
{{\phi _{WB}}}
\end{array}} \right].\]

Participants' state is a superposition of the six basis states
\begin{equation}
\begin{array}{l}
\left| \psi  \right\rangle  = {\psi _{AG}} \cdot \left| {{B_G}{A_A}} \right\rangle  + {\psi _{UG}} \cdot \left| {{B_G}{A_U}} \right\rangle  + {\psi _{WG}} \cdot \left| {{B_G}{A_W}} \right\rangle  + {\psi _{AB}} \cdot \left| {{B_B}{A_A}} \right\rangle \\
{\kern 1pt} {\kern 1pt} {\kern 1pt} {\kern 1pt} {\kern 1pt} {\kern 1pt} {\kern 1pt} {\kern 1pt} {\kern 1pt} {\kern 1pt} {\kern 1pt} {\kern 1pt} {\kern 1pt} {\kern 1pt} {\kern 1pt} {\kern 1pt}  + {\psi _{UB}} \cdot \left| {{B_B}{A_U}} \right\rangle  + {\psi _{WB}} \cdot \left| {{B_B}{A_W}} \right\rangle
\end{array}
\end{equation}
and the initial state corresponds to an amplitude distribution represented by the $6 \times {\rm{1}}$ column matrix
\[\psi \left( {\rm{0}} \right){\rm{ = }}\left[ {\begin{array}{*{20}{c}}
{{\psi _{AG}}}\\
{{\psi _{UG}}}\\
{{\psi _{WG}}}\\
{{\psi _{AB}}}\\
{{\psi _{UB}}}\\
{{\psi _{WB}}}
\end{array}} \right],\]
where, for example, ${\psi _{AG}}$ is the probability of observing state $\left| {{B_G}{A_A}} \right\rangle $. This probability distribution satisfies the constraint $L \cdot \phi  = 1$, where $L = \left[ {1,1,1,1,1,1} \right]$.
In this experiment, we assume that a face is shown to the participant at time zero. The initial probability distribution is uniformly random when the participant has no time to deliberate about the categorization and the actions to take, namely $\psi _{UG} = \psi _{UB} = 0.5$.

\textbf{Step 3: Inferences based on prior information}

During the decision process, new information at time ${t_1}$ changes the initial state ar time $t=0$ into a new state at time ${t_1}$.
In the C-D condition, if the face is categorized as good, the state changes to
\begin{equation}
\phi \left( {{t_1}} \right) = {\rm{ }}\frac{1}{{{\phi _{AG}} + {\rm{ }}{\phi _{UG}} + {\phi _{WG}}}}\left[ {\begin{array}{*{20}{c}}
{{\phi _{AG}}}\\
{{\phi _{UG}}}\\
{{\phi _{WG}}}\\
0\\
0\\
0
\end{array}} \right] = \left[ {\begin{array}{*{20}{c}}
{{\phi _G}}\\
{\textbf{0}}
\end{array}} \right].
\end{equation}
The initial probability that the face is categorized as good equals ${{\phi _{AG}} + {\rm{ }}{\phi _{UG}} + {\phi _{WG}}}$, and so the $3 \times 1$ matrix ${\phi _G}$ is the conditional probability distribution across actions given the categorization is good.

If the face is categorized as bad, the state changes to
\begin{equation}
\phi \left( {{t_1}} \right) = {\rm{ }}\frac{1}{{{\phi _{AB}} + {\rm{ }}{\phi _{UB}} + {\phi _{WB}}}}\left[ {\begin{array}{*{20}{c}}
0\\
0\\
0\\
{{\phi _{AB}}}\\
{{\phi _{UB}}}\\
{{\phi _{WB}}}
\end{array}} \right] = \left[ {\begin{array}{*{20}{c}}
{\textbf{0}}\\
{{\phi _B}}
\end{array}} \right].
\end{equation}
The initial probability that the face is categorized as bad equals ${{\phi _{AB}} + {\rm{ }}{\phi _{UB}} + {\phi _{WB}}}$, and so the $3 \times 1$ matrix ${\phi _B}$ is the conditional probability distribution across actions given the categorization is bad.

In the D-alone condition, the state remains the same as the initial state, which is a mixed state produced by a weighted average of the distribution for two known conditions in C-D condition:
\begin{equation}
\begin{array}{l}
\phi \left( {{t_1}} \right) = \phi \left( 0 \right)\\
{\kern 29pt} = \left[ {\begin{array}{*{20}{c}}
{\left( {{\phi _{AG}} + {\phi _{UG}} + {\phi _{WG}}} \right) \cdot {\phi _G}}\\
{\left( {{\phi _{AB}} + {\phi _{UB}} + {\phi _{WB}}} \right) \cdot {\phi _B}}
\end{array}} \right]\\
{\kern 29pt} = \left( {{\phi _{AG}} + {\phi _{UG}} + {\phi _{WG}}} \right) \cdot \left[ {\begin{array}{*{20}{c}}
{{\phi _G}}\\
{\bf{0}}
\end{array}} \right] + \left( {{\phi _{AB}} + {\phi _{UB}} + {\phi _{WB}}} \right) \cdot \left[ {\begin{array}{*{20}{c}}
{\bf{0}}\\
{{\phi _B}}
\end{array}} \right].
\end{array}
\end{equation}

\textbf{Step 4: Strategies based on payoffs}

During the decision making process, the participants need to evaluate the payoffs in order to select an appropriate action, which evolve the state at time ${t_1}$ into a new state at time ${t_2}$. The evolution of the state during this time period corresponds the thought process leading to a action decision, defection, cooperation or hesitance.
The evolution of the state obeys a Kolmogorov forward equation (Eq. (\ref{Kolmogorov})) driven by a $6 \times 6$ intensity matrix $K$. The solution is:
\[\phi \left( {{t_2}} \right) = {e^{ Kt}} \cdot \phi \left( {{t_1}} \right).\]
For $t = {t_2} - {t_1}$, the state to state transition matrix is defined by $T\left( t \right) = {e^{tK}}$ with ${T_{ij}}\left( t \right)$ represents the probability of transiting to state $i$ at ${t_2}$ given being in state $j$ at time ${t_2}$.
In this experiment, $t$ is set to 2 corresponding to the average time that a participant takes to make a decision (approximately 2 seconds)\cite{Busemeyer2009Empirical}.
The intensity matrix $K$ is
\begin{equation}
H = \left[ {\begin{array}{*{20}{c}}
{{K_G}}&0\\
0&{{K_B}}
\end{array}} \right]
\end{equation}
with
\[{K_G} = \left( {\begin{array}{*{20}{c}}
- \frac{{3{k_r} + {k_w}}}{2}&{k_r}&{k_r}\\
\frac{{{k_r} + {k_w}}}{2}&{-({k_r}+{k_w})}&\frac{{{k_r} + {k_w}}}{2}\\
{k_w}&{k_w}&- \frac{{{k_r} + 3{k_w}}}{2}
\end{array}} \right),\]
\[{K_B} = \left( {\begin{array}{*{20}{c}}
- \frac{{{k_r} + 3{k_w}}}{2}&{k_w}&{k_w}\\
\frac{{{k_r} + {k_w}}}{2}&{-({k_r}+{k_w})}&\frac{{{k_r} + {k_w}}}{2}\\
{k_r}&{k_r}&- \frac{{3{k_r} + {k_w}}}{2}
\end{array}} \right).\]
The $3 \times 3$ intensity matrix ${K_G}$ applies when the face is categorized as good, and the other matrix ${K_B}$ applies when the face is categorized as bad. The parameter ${k_r}$ represents the payoff for taking the right choice, namely attacking the bad guy and withdrawing the good guy, and the parameter ${k_r}$ represents the payoff for taking the wrong choice, namely attacking the good guy and withdrawing the bad guy. The off-diagonal elements of $K$ are nonnegative, and the columns of $K$ sum to zero, which then guarantees that the column of $T$ sum to one, which finally guarantees that $\phi \left( t \right) $.

In the C-D condition, if the face is categorized as good, the state changes to
\begin{equation}
\phi \left( {{t_2}} \right) = {e^{  K \cdot t}} \cdot \phi ({t_{\rm{1}}}){\rm{ = }}\left[ {\begin{array}{*{20}{c}}
{{e^{  {K_G} \cdot t}}}&0\\
0&{{e^{  {K_B} \cdot t}}}
\end{array}} \right] \cdot \left[ {\begin{array}{*{20}{c}}
{{\phi _G}}\\
\textbf{0}
\end{array}} \right] = {e^{  {K_G} \cdot t}} \cdot {\phi _G}
\end{equation}
If the face is categorized as bad, the state changes to
\begin{equation}
\phi \left( {{t_2}} \right) = {e^{  K \cdot t}} \cdot \phi ({t_{\rm{1}}}){\rm{ = }}\left[ {\begin{array}{*{20}{c}}
{{e^{ {K_G} \cdot t}}}&0\\
0&{{e^{ {K_B} \cdot t}}}
\end{array}} \right] \cdot \left[ {\begin{array}{*{20}{c}}
\textbf{0}\\
{{\phi _B}}
\end{array}} \right] = {e^{ {K_B} \cdot t}} \cdot {\phi _B}
\end{equation}
In the D-alone condition, the state changes to
\begin{equation}
\footnotesize
\begin{array}{*{20}{l}}
\begin{array}{l}
\phi \left( {{t_2}} \right) = {e^{K \cdot t}} \cdot \phi (0)\\
{\kern 23pt} = \left[ {\begin{array}{*{20}{c}}
{{e^{{K_G} \cdot t}}}&0\\
0&{{e^{{K_B} \cdot t}}}
\end{array}} \right] \cdot \left[ {\begin{array}{*{20}{c}}
{\left( {{\phi _{AG}} + {\phi _{UG}} + {\phi _{WG}}} \right) \cdot {\phi _G}}\\
{\left( {{\phi _{AB}} + {\phi _{UB}} + {\phi _{WB}}} \right) \cdot {\phi _B}}
\end{array}} \right]
\end{array}\\
{\kern 30pt}{ = \left( {{\phi _{AG}} + {\phi _{UG}} + {\phi _{WG}}} \right) \cdot {e^{{K_G} \cdot t}} \cdot {\phi _G}{\rm{ + }}\left( {{\phi _{AG}} + {\phi _{UG}} + {\phi _{WG}}} \right) \cdot {e^{{K_B} \cdot t}} \cdot {\phi _B}}.
\end{array}
\end{equation}

\textbf{Step 5: Basic probability assignments measurement}

According to section \ref{responsepro}, response probability of each decision can be derived. In an evidential framework, the BPA can also be derived. Specially, when the focal elements of a mass function is one, the BPA is the same with a probability. The measurement matrix is defined as
\[M = \left( {\begin{array}{*{20}{c}}
{{M_G}}&\textbf{0}\\
\textbf{0}&{{M_B}}
\end{array}} \right).\]
In the C-D condition, if the face is categorized as good, the $3 \times 3$ matrix ${M_G}$ is set to
\[{M_G} = \left[ {\begin{array}{*{20}{c}}
1&0&0\\
0&0&0\\
0&0&0
\end{array}} \right],\left[ {\begin{array}{*{20}{c}}
0&0&0\\
0&1&0\\
0&0&0
\end{array}} \right]or\left[ {\begin{array}{*{20}{c}}
0&0&0\\
0&0&0\\
0&0&1
\end{array}} \right]\]
respectively to pick out the state of attacking, hesitance or defecting while the $3 \times 3$ matrix ${M_B}$ is set to \textbf{0}.
If the face is categorized as bad, the $3 \times 3$ matrix ${M_B}$ is set to
\[{M_B} = \left[ {\begin{array}{*{20}{c}}
1&0&0\\
0&0&0\\
0&0&0
\end{array}} \right],\left[ {\begin{array}{*{20}{c}}
0&0&0\\
0&1&0\\
0&0&0
\end{array}} \right]or\left[ {\begin{array}{*{20}{c}}
0&0&0\\
0&0&0\\
0&0&1
\end{array}} \right]\]
respectively to pick out the state of attacking, hesitance or defecting while the $3 \times 3$ matrix ${M_G}$ is set to \textbf{0}.
In the D-alone condition the $3 \times 3$ matrixes ${M_B}$ and ${M_G}$ are set to
\[{M_B} = {M_G} = \left[ {\begin{array}{*{20}{c}}
1&0&0\\
0&0&0\\
0&0&0
\end{array}} \right],\left[ {\begin{array}{*{20}{c}}
0&0&0\\
0&1&0\\
0&0&0
\end{array}} \right]or\left[ {\begin{array}{*{20}{c}}
0&0&0\\
0&0&0\\
0&0&1
\end{array}} \right]\]
respectively to pick out the state of attacking, hesitance or defecting.

Using $L = \left[ {1,1,1,1,1,1} \right]$, the BPA of each state equals
\begin{equation}
m = L \cdot M \cdot {e^{tK}} \cdot \phi \left( {{t_1}} \right).
\end{equation}
The derived BPAs consist body of evidences, $\left( {{\Re _{CD}},{m_{CD}}} \right)$ and $\left( {{\Re _D},{m_D}} \right)$ for two conditions respectively.
\[\begin{array}{l}
\left( {{\Re _{CD}},{m_{CD}}} \right) = \left[ {m\left( {AG} \right),m\left( {UG} \right),m\left( {WG} \right),m\left( {AB} \right),m\left( {UB} \right),m\left( {WG} \right)} \right]\\
{\kern 62pt} = \left( {0.0264,0.0811,0.0625,0.3050,0.3960,0.1290} \right)
\end{array}\]
\[\left( {{\Re _D},{m_D}} \right) = \left[ {m\left( AU \right),m\left( UU \right),m\left( WU \right)} \right] = \left( {0.3314,0.4771,0.1915} \right)\]

\textbf{Step 6: Probability distribution based on uncertainty measurement}

To obtain the probability distribution, the BPA of uncertain state should be transferred. It is still an open issue to address it. The classical method is using pignistic probability transformation. Obviously, however, the disjunction effect can not be predicted in this way.
To address it, a new parameter $\gamma$ which represents the extra uncertainty degree (EUD) is determined as follows:
\begin{equation}
\gamma {\rm{ = }}\left| {\frac{{{E_D} - {E_{CD}}}}{{{E_D} + {E_{CD}}}}} \right|,
\end{equation}
where ${E_{CD}}$ and ${E_D}$ are the information volume of $\left( {{\Re _{CD}},{m_{CD}}} \right)$ and $\left( {{\Re _D},{m_D}} \right)$ respectively calculated by Deng entropy as follows:
\[\begin{array}{l}
{E_D} =  - 0.0264{\log _2}\left( {\frac{{0.0264}}{{{2^1} - 1}}} \right) - 0.0811{\log _2}\left( {\frac{{0.0.0811}}{{{2^2} - 1}}} \right) - 0.0625{\log _2}\left( {\frac{{0.0625}}{{{2^1} - 1}}} \right)\\
{\kern 17pt} - 0.3050{\log _2}\left( {\frac{{0.3050}}{{{2^1} - 1}}} \right) - 0.3960{\log _2}\left( {\frac{{0.3960}}{{{2^2} - 1}}} \right) - 0.1290{\log _2}\left( {\frac{{0.1290}}{{{2^1} - 1}}} \right)\\
{\kern 17pt} = 2.8715
\end{array},\]
\[\begin{array}{l}
{E_D} =  - 0.3314{\log _2}\left( {\frac{{0.3314}}{{{2^2} - 1}}} \right) - 0.4771{\log _2}\left( {\frac{{0.4771}}{{{2^4} - 1}}} \right) - 0.1915{\log _2}\left( {\frac{{0.1915}}{{{2^2} - 1}}} \right)\\
{\kern 17pt} = 4.1868.
\end{array}\]
The cardinality of states $AG$, $WG$, $AB$, $WB$ is one, the cardinality of states $UG$, $UB$, $AU$, $WU$ is two, and the cardinality of the state $UU$ is 4.

Without a categorization, the uncertainty for the D-alone condition is obviously larger than the C-D condition. It is reasonable to assume that the hesitated participants are tougher to take a final decision, especially for the narrow type faces.
According to the cognitive dissonance and social projection in psychology, people tend to be consistent with their beliefs and actions\cite{Festinger1962A,Krueger2012Social,busemeyer2012social}. The hesitated people are more tend to attack especially in the D-alone condition.
Hence, for $\left( {{\Re _D},{m_D}} \right)$, the BPA of the uncertain state should be assigned more to the attacking based on the EUD.
Then the probability of attacking in the D-alone condition is calculated as
\begin{equation}\label{Pa_D}
{P_D}\left( A \right) = m(AU) + \left( {\frac{1}{2} + \gamma } \right)m\left( {UU} \right)=0.6589.
\end{equation}
As the observed experiments result shows that the disjunction effect is less than 0.1, the information volume will not differ hugely, which guarantees $\gamma  < 0.5$.
For $\left( {{\Re _{CD}},{m_{CD}}} \right)$, the BPA of uncertain state is still handled by using PPT (Eq. (\ref{PPT})).
Then the probability of attacking in the C-D condition is calculated as
\begin{equation}\label{Pa_CD}
{P_{CD}}\left( A \right) = m\left( {AG} \right) + \frac{1}{2}m\left( {UG} \right) + m\left( {AB} \right) + \frac{1}{2}m\left( {UB} \right)= 0.57
\end{equation}
The difference value of Eq. (\ref{Pa_D}) and Eq. (\ref{Pa_CD}) is the predicted disjunction effect.
\[Dis = {P_D}\left( A \right) - {P_{CD}}\left( A \right)=0.0889\]
According to the above steps, the comprehensive process of our model is illustrated in Figure \ref{process}.

\section{Model results and comparisons}
\subsection{Comparison with experiment results}
Applying the EM model to the categorization decision-making experiments, the probability distributions for all the experiments can be obtained.
Compare the obtained model results with the observed experiment results (for narrow type face), the model results are close to the practical situation, which verifies the correctness and effectiveness of our model.
As Table \ref{result compare} shows, the disjunction effect is well predicted and the average error rate is only approximately 0.5\%, which proof the correctness and effectiveness of our model greatly.
\begin{table}[!h]
\centering
\scriptsize
\caption{The results of the EM model}
\label{result compare}
\begin{threeparttable}
\begin{tabular}{ccccccccc}
\toprule
Literature                                                                                        &     & $P\left( G \right)$ & $P\left( {A|G} \right)$\tnote{1} & $P\left( B \right)$ & $P\left( {A|B} \right)$\tnote{2} & ${{P}_T}$     & $P\left( A \right)$  &$Dis$  \\
\midrule
\multirow{2}{*}{\scriptsize\begin{tabular}[c]{@{}c@{}}Townsend \\ $et\ al.$ (2000)\cite{Townsend2000Exploring}\end{tabular}}                  & Obs & 0.17 & 0.41   & 0.83 & 0.63   & 0.59   & 0.69  & 0.1 \\
                                                                                                               & EM & 0.17 & 0.394   & 0.83 & 0.606 & 0.57 & 0.6589 &0.0889\\ \hline
\multirow{2}{*}{\scriptsize\begin{tabular}[c]{@{}c@{}}Busemeyer \\ $et\ al.$ (2009)\cite{Busemeyer2009Empirical}\end{tabular}}                 & Obs & 0.20 & 0.45   & 0.80 & 0.64   & 0.60   & 0.69  &0.09 \\
                                                                                                               & EM & 0.20  & 0.4019 & 0.5981 & 0.5588 &0.80 & 0.6404     & 0.0816 \\ \hline
\multirow{2}{*}{\scriptsize\begin{tabular}[c]{@{}c@{}}Wang \& Busemeyer (2016)\\  experiment 1\cite{wang2016interference}\end{tabular}} & Obs & 0.21 & 0.41     & 0.79 & 0.58   & 0.54   & 0.59  &0.05 \\
                                                                                                             & EM & 0.21 & 0.3840   & 0.79 & 0.6160 & 0.5673 & 0.6432 &0.0759\\
                                                                                                             \hline
\multirow{2}{*}{\scriptsize\begin{tabular}[c]{@{}c@{}}Wang \& Busemeyer (2016) \\ experiment 2\cite{wang2016interference}\end{tabular}} & Obs & 0.24 & 0.37   & 0.76 & 0.61   & 0.55   & 0.60   &0.05\\
                                                                                                             & EM & 0.24 & 0.3384 & 0.76 & 0.6197 & 0.5622 & 0.63 &0.0678\\
                                                                                                             \hline
\multirow{2}{*}{\scriptsize\begin{tabular}[c]{@{}c@{}}Wang \& Busemeyer (2016) \\ experiment 3\cite{wang2016interference}\end{tabular}} & Obs & 0.24 & 0.33   & 0.76 & 0.66   & 0.58   & 0.62  &0.04 \\
                                                                                                             & EM & 0.24 & 0.3384 & 0.76 & 0.6616 & 0.5841 & 0.6436
                                                                                                             &0.0596\\
                                                                                                  \hline
\multirow{2}{*}{Average}                                                                          & Obs & 0.21 & 0.39   & 0.79 & 0.62   & 0.57   & 0.64   &0.07\\
                                                                                                  & EM & 0.21 & 0.3797 & 0.79 & 0.6203 & 0.5685 & 0.6432 &0.0747\\
\bottomrule
\end{tabular}
 \begin{tablenotes}
        \footnotesize
        \item[1] $Dis$ represents the disjunction effect.
        \item[2] Obs represents the observed experiment results.
        \item[3] EM represents the results of the Evidential Markov model.
      \end{tablenotes}
\end{threeparttable}
\end{table}
\subsection{Comparison with other uncertainty measurement method}
\begin{figure}[!ht]
\centering
\includegraphics[scale=0.4]{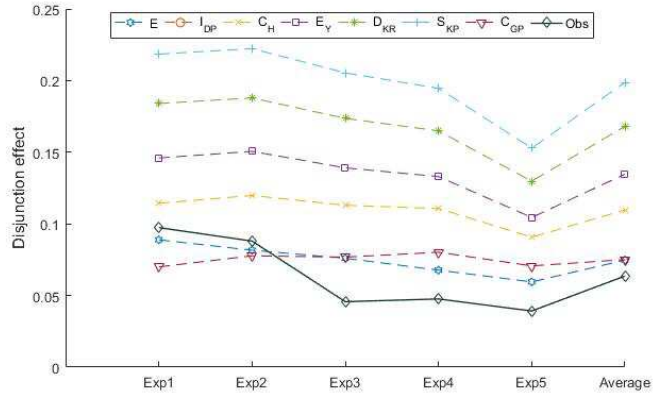}
\caption{The predicted disjunction effect of different methods}\label{disjunction}
\end{figure}

\begin{figure}[!ht]
\centering
\includegraphics[scale=0.4]{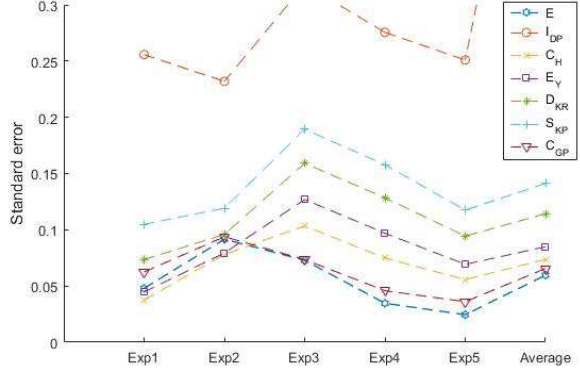}
\caption{The SE of different methods}\label{standard}
\end{figure}

The crucial step of the EM model to predict the disjunction effect is the measurement of the extra uncertainty. Deng entropy is compared with many other uncertainty measuring methods (shown in Table \ref{uncertainty measure}) in Deng (2016)\cite{dengentropy}. These methods are still used to do the comparison in our method.
\begin{table}[!ht]
\centering
\scriptsize
\caption{Methods of measuring uncertainty}
\label{uncertainty measure}
\begin{tabular}{ll}
\toprule
Name & Formula \\
\midrule
Deng entropy\cite{dengentropy} & $E\left( m \right) =  - \sum\limits_{A \in X} {m\left( A \right){{\log }_2}\frac{{m\left( A \right)}}{{{2^{\left| A \right|}} - 1}}} $             \\
Dubois \& Prade¡¯s weighted Hartley entropy\cite{DIDIER1984A}    &   ${I_{DP}}\left( m \right) =  - \sum\limits_{A \in X} {m\left( A \right){{\log }_2}\left| A \right|} $      \\
Hohle¡¯s confusion measure\cite{hohle1982entropy}    &  ${C_H}\left( m \right) =  - \sum\limits_{A \in X} {m\left( A \right){{\log }_2}Bel\left( A \right)} $ \\
Yager¡¯s dissonance measure\cite{RONALD1982ENTROPY}   &  ${E_Y}\left( m \right) =  - \sum\limits_{A \in X} {m\left( A \right){{\log }_2}Pl\left( A \right)} $  \\
Klir \& Ramer¡¯s discord\cite{GEORGE1990UNCERTAINTY}    &   ${D_{KR}}\left( m \right) =  - \sum\limits_{A \in X} {m\left( A \right){{\log }_2}\sum\limits_{B \in X} {m\left( B \right)\frac{{\left| {A \cap B} \right|}}{{\left| B \right|}}} } $      \\
Klir \& Parviz¡¯s strife\cite{Klir1992138}    &    ${S_{KP}}\left( m \right) =  - \sum\limits_{A \in X} {m\left( A \right){{\log }_2}\sum\limits_{B \in X} {m\left( B \right)\frac{{\left| {A \cap B} \right|}}{{\left| B \right|}}} } $     \\
George \& Pal¡¯s conflict measure\cite{THOMAS1996QUANTIFICATION}    & ${C_{GP}}\left( m \right) = \sum\limits_{A \in X} {m\left( A \right){{\log }_2}\sum\limits_{B \in X} {m\left( B \right)\left[ {1 - \frac{{\left| {A \cap B} \right|}}{{\left| {A \cup B} \right|}}} \right]} } $\\
\bottomrule
\end{tabular}
\end{table}
The comparison of the predicted disjunction effect and the standard error (SE) is illustrated in Figure \ref{disjunction} and Figure \ref{standard} respectively, where the Exp 1 to 5 comes from Townsend $et\ al.$ (2000), Busemeyer $et\ al.$ (2009), Wang \& Busemeyer (2016) experiment 1, 2, 3 respectively. We can see that the SE of Deng entropy is small for each experiment and the average SE of Deng entropy is the smallest; the predicted disjunction effect of Deng entropy is close to the observed one for each experiment and the average disjunction effect is the most accurate. Deng entropy has a powerful ability of measuring the uncertainty of a BOE, especially when many focal sets are multi-element. Hence, Deng entropy is used to measure the extra uncertainty degree in the EM model.
\subsection{Comparison with other models}
As illustrated in section \ref{LawinM}, the classical Markov model can not predict the disjunction effect. However, many quantum models have been proposed to explain the disjunction fallacy. In a quantum framework, before taking a decision, human thoughts are seen as superposed waves that can interfere with each other, influencing the final decision.
As shown in Figure \ref{quantumprocess}, different decisions can coexist at one time before the final decision is made, while the Markov model assumes that the decision at one time must be certain.
The idea of the EM model is identified with the quantum model in some degree, namely allowing a state to be uncertain.

The interference effect, a term of quantum mechanics, is borrowed to account for the disjunction fallacy.
One of the effective quantum models is the quantum dynamical (QD) model first proposed by Busemeyer $et\ al.$ in 2006\cite{Busemeyer2006Quantum}, which is always used to compare with the classical Markov model.
Busemeyer $et\ al.$ (2009)\cite{Busemeyer2009Empirical} illustrates that the QD model can predict the interference effect while the Markov model can not in detail.
Both the QD model and the Markov model are formulated as a random walk decision process, but the QD model describes the evolution of complex valued probability amplitudes over time. In a quantum framework, the state evolves obeying a Schr${\ddot o}$dinger equation (Eq. \ref{schrodinger}), which is driven by a Hermitian matrix. Also, two payoffs functions are used to fill the Hermitian matrix $H$.
\begin{equation}\label{schrodinger}
\frac{d}{{dt}}\psi \left( t \right) =  - i \cdot H \cdot \psi \left( t \right)
\end{equation}
Admittedly, the disjunction effect can be predicted in the QD model, however, its problem is that the model requires a growth of parameters. For example, in Wang \& Busemeyer (2016)\cite{wang2016interference}, a modified QD model, the quantum belief-action entanglement (BAE) model, is proposed. The actions and beliefs are deemed to be entangled in some degree, which is the key to produce the interference effect. The entanglement degree is described in a introduced parameter whose value is set artificially.
However, in our EM model, the extra uncertainty degree is totally driven by data. And it turns out that the uncertainty can be well measured in our model.
To see the ability of predicting the disjunction effect, the model results of the probability of attacking for the D-alone condition is compared as shown in Figure \ref{zhuzhuangtu}).

As we can see, both the QD model and the EM model can predict the disjunction effect while the Markov model can not.
However, the prediction result of our EM model is a bit more accurate.
Also the free parameter of EM model is less.
Based on the above, it is reasonable to conclude that our model is correct and efficient.
\section{Conclusion}\label{Conclusion}
To model the decision making process and explain the disjunction fallacy, the EM decision making model is proposed in this paper.
The model combines Dempster-Shafer evidence theory with the Markov model.
The classical Markov model assumes that the at a certain time, people's decision has to be certain at one time.
In an evidential framework, the action states are extended.
The hesitance of a decision maker is represented by a new introduced uncertain state.
Hence, the state is allowed to be uncertain at one time before a final decision is made.
The uncertainty measurement is the key to predict the disjunction effect.
A belief entropy, named Deng entropy, is used due to its powerful ability of uncertainty measurement.
The model results show that the EM model can well predict the disjunction effect, which is impossible for the classical Markov model.
Also, the EM model has less free parameters than the quantum model.
Hence, it is reasonable for us to conclude that the EM model is correct and effective.
\section*{Acknowledgement}
The work is partially supported by National Natural Science Foundation of China (Grant No. 61671384), Natural Science Basic Research Plan in Shaanxi Province of China (Program No. 2016JM6018), Aviation Science Foundation (Program No. 20165553036), the Fund of SAST (Program No. SAST2016083)

\bibliographystyle{model1-num-names}
\bibliography{myreference}

\end{document}